\documentclass[11pt]{article}

\usepackage[final]{acl}

\usepackage{times}
\usepackage{latexsym}
\usepackage{enumitem}

\usepackage[T1]{fontenc}

\usepackage[utf8]{inputenc}

\usepackage{microtype}

\usepackage{inconsolata}

\usepackage{graphicx}

\usepackage{algorithm}
\usepackage{algpseudocode}
\usepackage{amsmath}
\usepackage{booktabs}
\usepackage{float}
\usepackage{titletoc}
\usepackage{etoc}

\usepackage[most]{tcolorbox}
\usepackage{tabularx}

\title{When LLMs Stop Following Steps: A Diagnostic Study of Arithmetic Procedural Execution in Language Models}

\author{
\textbf{Sailesh Panda}$^{\dagger}$, 
\textbf{Pritam Kadasi}$^{\dagger}$, 
\textbf{Abhishek Upperwal}$^{\ddagger}$, 
\textbf{Mayank Singh}$^{\dagger}$ \\
\\
$^{\dagger}$Indian Institute of Technology Gandhinagar \\
$^{\ddagger}$Soket AI \\
\\
\small{\textbf{Correspondence:} \href{mailto:sailesh.panda@iitgn.ac.in}{sailesh.panda@iitgn.ac.in}}
}

\usepackage{float}
\usepackage{graphicx}
\usepackage{subcaption}
\usepackage{geometry}

\usepackage{listings}
\usepackage{xcolor}

\newif\ifrevision
\revisionfalse
\newcommand{\revcolor}{\ifrevision\color{blue}\fi}
\newcommand{\rev}[1]{{\revcolor #1}}

\begin{document}
\maketitle


\begin{abstract}
Large language models (LLMs) often achieve strong performance on reasoning benchmarks, but final-answer accuracy alone does not show whether they faithfully execute the procedure specified in a prompt. We introduce a controlled diagnostic benchmark for \emph{arithmetic procedural execution}, where models are given a step-wise arithmetic procedure and two numeric inputs, and must return the final computed value. Complexity is varied through procedure length and look-back dependencies over intermediate variables. Average first-answer accuracy drops from \textbf{63\%} on 5-step procedures to \textbf{20\%} on 95-step procedures. Generation-level analysis shows that failures often involve missing answers and under-executed traces, and, for models that show their steps, step-level verification shows that the share of correctly executed steps before the first error falls from 89\% to 28\%. These findings reveal a consistent decline in execution performance as arithmetic procedural complexity increases.
\end{abstract}
\section{Introduction}
\label{intro}

\label{intro}

\label{intro}

Large language models (LLMs) show strong performance on many reasoning tasks, including arithmetic, symbolic manipulation, multi-step question answering, and code generation \citep{wei2022chain, wei2022emergent, weng-etal-2023-large, zhao2025sample, zhou2022teaching}. Recent progress is driven by scale, instruction tuning, and reinforcement learning methods that encourage longer reasoning traces \citep{guo2025deepseek, shao2024deepseekmath}. 

However, strong final-answer performance does not necessarily reflect faithful execution of the intended procedure. This is critical for procedure-following tasks, where models must execute specified multi-step procedures, track intermediate states, apply operations in order, and terminate correctly. This matters for agent workflows and tool chains~\citep{lu-etal-2025-toolsandbox}, where a model executes an explicit plan. There, a skipped or wrong step can silently corrupt the result. Prior work shows LLMs often rely on shortcut heuristics, shallow patterns, or unstable computations, failing even when the complete procedure is provided~\citep{elazar2022measuring, shojaee2025illusion, yang-etal-2025-unveiling, xu-etal-2025-principled}. Yet, existing studies often focus on broad benchmarks or final answers, making it difficult to isolate how performance changes when procedures are fully specified, operations are simple, and complexity scales primarily with length.
We study \emph{arithmetic procedural execution}: an LLM's ability to carry out an explicitly specified arithmetic procedure and return the correct final value. We introduce a controlled diagnostic benchmark where each example provides a step-wise arithmetic procedure and two numeric inputs. The procedure defines intermediate variables using basic operations, requiring the model to execute steps in order to compute the final value. Because the reference output is deterministic, this setup minimizes ambiguity, allowing procedural execution to be studied under controlled conditions.

Our benchmark increases complexity along two axes: (1) \textbf{procedure length}, testing if accuracy holds as required steps increase, and (2) \textbf{look-back dependencies}, where a step requires retrieving intermediate variables from several steps earlier. This stresses procedural state tracking beyond elementary arithmetic. Across various input ranges, data types, operations, and look-back settings, the benchmark comprises 55 synthetic datasets totaling 55{,}000 examples (Section~\ref{sec:dataset_construction}). We evaluate 15 language models across multiple scales, reporting answer-level and generation-level metrics, including first-answer accuracy (FAA), Correct@Any (CAA), answer position, step-following behavior, and step-level verification (SLV). Our study is diagnostic: it quantifies how a known failure depends on procedure length and dependency depth, without proposing mechanisms or interventions.

We address two research questions:
\begin{itemize}[noitemsep,nosep]
    \item \textbf{RQ1:} How does model accuracy change as procedures lengthen and require deeper look-back dependencies?
    \item \textbf{RQ2:} What generation-level failure modes emerge during execution, and how do they relate to final-answer accuracy?
\end{itemize}

Empirically, average FAA decreases from \textbf{63\%} at 5 steps to \textbf{20\%} at 95 steps. Increasing dependency depth from the immediately preceding variable to up to seven steps back further reduces average accuracy by \textbf{23.85} percentage points. This indicates models struggle not only with length but also with retrieving and combining non-local intermediate states. Our contributions are: (1) a controlled, CC-BY 4.0 licensed synthetic benchmark for evaluating long-horizon arithmetic procedural execution;\footnote{\url{https://github.com/lingo-iitgn/arithmetic-procedural-reasoning}} (2) a comprehensive evaluation of 15 models (55{,}000 examples) analyzing the effects of procedure length and look-back dependency; and (3) generation-level and step-level analyses showing that failures often coincide with missing answers, under-execution, and inconsistent state tracking, suggesting that arithmetic errors alone may not fully explain the observed degradation.

\section{Related Work}
\label{sec:related_work}

LLM reasoning has been improved through prompting, scaling, self-verification, and reinforcement learning-based training methods \citep{wei2022chain, wei2022emergent, zhou2022teaching, weng-etal-2023-large, guo2025deepseek, shao2024deepseekmath}. We focus on work most related to procedural execution along four directions.

\subsection{Algorithmic Execution}
Prior work has studied whether neural models can learn and execute algorithmic procedures such as sorting, graph traversal, arithmetic, and program-like computation \citep{velickovic2021neural, xu2020what, schwarzschild2021can, weiss2021thinking, zhou2024what, pmlr-v202-abbe23a, kazemnejad2023the, yang2024aqa}. These studies often focus on whether models can learn an algorithm from data and generalize to longer or harder inputs. More recent work evaluates whether reasoning-oriented LLMs can execute explicitly specified procedures at inference time. \citet{shojaee2025illusion} show that models can fail on controllable puzzle tasks even when the full solution procedure is provided, while follow-up work finds that performance varies substantially with task design and complexity \cite{varela2025rethinking}. Our work follows this line but studies a simpler and more controlled setting: arithmetic procedures with deterministic ground truth and systematically varied execution length and dependency depth. Unlike puzzle studies, where problem size increases solution length and state size together, and perturbation benchmarks that pose word problems without an explicit procedure \citep{mirzadeh2025gsmsymbolic, zhou2025gsminfty}, we state every step, vary length and look-back depth independently, and verify each intermediate state. Our contribution is thus a controlled measurement of a known phenomenon rather than a new failure mode (Table~\ref{tab:prior_comparison}, Appendix~\ref{app:prior_comparison}).

\subsection{Synthetic Diagnostics}
Synthetic diagnostic benchmarks provide controlled settings for testing specific reasoning abilities, such as question answering, compositional instruction following, relational reasoning, and semantic parsing \citep{weston2015aicompletequestionansweringset, pmlr-v80-lake18a, sinha-etal-2019-clutrr, kim-linzen-2020-cogs}. Recent diagnostic studies further show that strong benchmark performance can hide brittle reasoning behavior under small perturbations or increasing complexity \citep{mirzadeh2025gsmsymbolic, zhou2025gsminfty, kohli-etal-2025-groundcocoa}. Our benchmark has a similar diagnostic goal, but focuses specifically on arithmetic procedural execution: the model is given the full procedure in the prompt, and complexity is controlled through procedure length, input type, operation type, and look-back dependency.

\subsection{State Tracking}
Many reasoning tasks require models to maintain and update intermediate states. Prior work has studied state tracking in procedural text, action-state reasoning, dialogue, and other structured settings \citep{dalvi-etal-2018-tracking, tandon-etal-2018-reasoning, li-etal-2021-implicit}. Recent work also examines whether LLMs encode intermediate reasoning states or rely on shortcut associations instead \citep{yang-etal-2025-unveiling}. Our task provides a controlled form of state tracking: procedures define intermediate variables $S_1, S_2, S_3, \ldots$, and later steps may depend on variables produced several steps earlier. By varying look-back depth, we test how well models retrieve and combine non-local intermediate states during execution.

\subsection{Compositional Generalization}
Compositional generalization studies whether models can recombine familiar parts in new or harder configurations \citep{pmlr-v80-lake18a, kim-linzen-2020-cogs, keysers2020measuring, hupkes2020compositionality}. For Transformers, prior work shows that length generalization and arithmetic generalization depend strongly on task structure and architectural choices such as positional encoding \citep{zhou2024what, kazemnejad2023the, xu-etal-2025-principled, huang2026transformers}. Our work is related because long procedures require repeated composition of simple operations. However, unlike training-time generalization studies, we test instruction-tuned and reasoning-oriented LLMs at inference time, where the full procedure is explicitly provided. Failures in this setting therefore point to weaknesses in execution, state tracking, or generation behavior rather than lack of access to the rule.

\section{Task Formulation}
\label{sec:task_formulation}

We study whether large language models can faithfully execute an explicitly specified multi-step procedure when the full procedure is provided in the prompt. To isolate procedural execution from broader linguistic ambiguity, we consider a controlled arithmetic setting.

Each example consists of a step-wise natural-language procedure and two numeric inputs, \(x\) and \(y\). The procedure initializes \(S_1=x\) and \(S_2=y\), then defines subsequent variables through arithmetic operations over previously computed variables. The model must execute the procedure exactly as written and return only the final numeric result. We vary procedure length to control task difficulty. Formally, for each step \(t \geq 3\),

\[
S_t = S_i \; o_t \; S_j,
\]

where \(o_t \in \{+, -, \times, \div\}\) denotes the operation at step \(t\), and \(S_i, S_j\) are selected from a bounded history of previously computed variables. The size of this accessible history is controlled by the look-back parameter \(k\), which determines how many earlier intermediate variables remain available for future computation. The final target output is the value computed in the last step of the procedure. Figure~\ref{fig:stepwise_example} shows a representative example of a generated procedure.

This setup gives us full control over arithmetic procedural complexity while keeping the underlying computation simple and verifiable. Since the correct answer is deterministically computable, the benchmark allows us to compare final-answer success with observable procedure-execution behavior.

\begin{figure}[t]
\centering

\begin{minipage}{0.95\linewidth}
\begin{lstlisting}[escapeinside={(*@}{@*)}]
function(x, y):
    Let (*@$S_1$@*) = x
    Let (*@$S_2$@*) = y
    Step 1: (*@$S_3$@*) = (*@$S_1$@*) + (*@$S_2$@*)
    Step 2: (*@$S_4$@*) = (*@$S_3$@*) - (*@$S_2$@*)
    Step 3: (*@$S_5$@*) = (*@$S_4$@*) / (*@$S_2$@*)
    Step 4: (*@$S_6$@*) = (*@$S_5$@*) * (*@$S_3$@*)
    Step 5: (*@$S_7$@*) = (*@$S_6$@*) - (*@$S_4$@*)
    Final Step: Return (*@$S_7$@*)
\end{lstlisting}
\end{minipage}

\caption{Representative step-wise arithmetic procedure. Later steps require retrieving and using earlier intermediate variables, making the task a test of procedural state tracking. For example, Step 5 computes \(S_7\) using \(S_6\) and \(S_4\), where \(S_4\) was computed three steps earlier (look-back 3).}
\label{fig:stepwise_example}

\end{figure}
\section{Experimental Setup}
\label{sec:experimental_setup}


We describe the dataset construction, evaluated models, inference procedure, and evaluation metrics used to study arithmetic procedural execution under increasing complexity.

\subsection{Dataset Construction}
\label{sec:dataset_construction}

We construct a controlled benchmark of step-wise arithmetic execution tasks following Section~\ref{sec:task_formulation}. Each instance consists of two numeric inputs ($x, y$), an explicit arithmetic procedure, and a deterministic reference output. We restrict the benchmark to two inputs to isolate procedural execution and state tracking from variable-management overhead. Models must execute the steps exactly and return the final value. By keeping elementary operations simple and scaling procedural burden via length and dependency structure, we evaluate sustained instruction execution without answer ambiguity.



\paragraph{Input ranges and data types.}
We generate inputs from three numeric ranges: $[0,1]$, $[1,10]$, and $[10,100]$. We use floating-point inputs for $[0,1]$, and both integer and floating-point for $[1,10]$ and $[10,100]$, yielding five input settings. Inputs are sampled uniformly: integers from discrete intervals (with replacement), and floats from continuous intervals (rounded to three decimal places).


\paragraph{Operation variants.}
We construct single-operation and mixed-operation procedures. Single-operation variants use only addition, subtraction, multiplication, or division. Mixed-operation procedures sample operators uniformly from all four at each step, requiring operator switching. This yields five operation variants per input setting (four single, one mixed), allowing us to compare homogeneous and heterogeneous sequences.


\paragraph{Procedure Horizons.}
We vary procedure length from 5 to 95 steps in increments of 10 to study longer horizons while maintaining computational tractability. We generate 100 examples per horizon, yielding 1{,}000 examples per dataset.


\paragraph{Look-back dependencies.}
To stress intermediate state tracking, we introduce look-back dependencies. For a look-back depth $k$, the accessible intermediate variables at step $t$ are:
\[
\mathcal{H}_t^{(k)} = \{S_2\} \cup \{S_j \mid \max(3,t-k) \leq j < t \}
\]
Here $t \geq 4$, since $S_3 = S_1 + S_2$. $S_2$ stays available at every step, so two distinct operands always exist.
Two distinct operands, $S_i, S_j \sim \mathcal{H}_t^{(k)}$, are sampled uniformly without replacement, and the next state is computed as:
\[
S_t = S_i \; o_t \; S_j, \quad \text{where } o_t \in \{+, -, \times, \div\}.
\]
We evaluate $k \in \{1, \dots, 7\}$ to study progressively longer, non-local dependency chains while maintaining computational tractability and deterministic ground-truth computation.

In total, the benchmark comprises 55 datasets and 55{,}000 examples.\footnote{We omit the full factorial design to manage experimental cost. For \textbf{look-back 1}, we include all five input settings and five operation variants ($5 \times 5 \times 1 = 25$ datasets). For higher look-backs, we restrict to the mixed-operation variant across six additional configurations ($5 \times 1 \times 6 = 30$ datasets). With 1{,}000 examples per dataset, the total is $(25+30)\times 1{,}000 = 55{,}000$ examples.}

\paragraph{Reference outputs.}
We compute reference outputs using an external deterministic executor that parses the procedure, executes each step, and returns the intermediate and final values. We intentionally retain edge cases (e.g., division-by-zero, numerical instabilities, infinities, NaNs) to evaluate if models correctly propagate or report them. The executor implementation is in Appendix~\ref{app:expected_output}.


\subsection{Model Selection}
\label{sec:model_selection}


We evaluate a diverse set of language models spanning different scales, model families, and reasoning-oriented training procedures. The selected models range from small models with approximately 1.5B parameters to large-scale frontier and mixture-of-experts models. For analysis, we group the models into five parameter-scale categories: small ($\leq$4B), mid-scale (7B), large (14B), very large (24B--30B), and frontier-scale ($\geq$100B). The evaluated models are listed in Table~\ref{tab:models}.

\begin{table}[t]
\centering
\scriptsize
\caption{Language models evaluated in our experiments.}
\label{tab:models}

\begin{tabularx}{\columnwidth}{lXc}
\toprule
\textbf{Abbrev.} & \textbf{Model} & \textbf{Params} \\
\midrule

\multicolumn{3}{l}{\textit{Small models ($\leq$4B)}} \\
DS-1.5 & DeepSeek-R1-Distill-Qwen-1.5B~\cite{guo2025deepseek} & 1.5B \\
QW-4 & Qwen3-4B-Thinking-2507~\cite{yang2025qwen3} & 4B \\

\midrule
\multicolumn{3}{l}{\textit{Mid-scale models (7B)}} \\
OL-7 & Olmo-3-7B-Think~\cite{olmo2025olmo} & 7B \\
MI-7 & Mistral-7B-Instruct-v0.3~\cite{mistral7b_instruct_v03} & 7B \\
DS-7 & DeepSeek-R1-Distill-Qwen-7B~\cite{guo2025deepseek} & 7B \\
OT-7 & OpenThinker2-7B~\cite{guha2025openthoughts} & 7B \\

\midrule
\multicolumn{3}{l}{\textit{Large models (14B)}} \\
MN-14 & Ministral-3-14B-Reasoning-2512~\cite{liu2026ministral} & 14B \\
DS-14 & DeepSeek-R1-Distill-Qwen-14B~\cite{guo2025deepseek} & 14B \\
QW-14 & Qwen3-14B~\cite{yang2025qwen3} & 14B \\

\midrule
\multicolumn{3}{l}{\textit{Very large models (24B--30B)}} \\
MG-24 & Magistral-Small-2509~\cite{rastogi2025magistral} & 24B \\
SV-30 & Sarvam-30B~\cite{sarvam_sovereign_models} & 30B \\
NM-30 & NVIDIA-Nemotron-3-Nano-30B~\cite{blakeman2025nvidia} & 30B \\

\midrule
\multicolumn{3}{l}{\textit{Frontier-scale models ($\geq$100B)}} \\
OA-120 & gpt-oss-120b~\cite{agarwal2025gpt} & 120B \\
DS-685 & DeepSeek-V3.2~\cite{deepseekai2025deepseekv32} & 685B \\
KK-1T & Kimi-K2.5~\cite{team2026kimi} & 1.1T \\

\bottomrule
\end{tabularx}
\end{table}

\subsection{Inference Procedure}
\label{sec:inference_procedure}

For each model and benchmark variant, we perform inference on all examples using a fixed prompt format. The prompt contains the system instruction, the full step-wise procedure, and the two input values. The model is instructed to execute the procedure exactly and return the final numeric answer enclosed within \texttt{<answer>} and \texttt{</answer>} tags. The prompt asks for both the intermediate steps and a tagged answer; we score the first \texttt{<answer>} span (Appendix~\ref{sec:prompt}).

After generation, we extract all answer spans matching the required tag format. If the model produces multiple tagged answers, we record both the first extracted answer and the set of all extracted answers. If the output contains no valid tagged answer, the prediction is marked as \textsc{null}. All extracted answers are compared against the deterministic reference output after rounding to three decimal places.

All experiments were conducted on NVIDIA H200 (141 GB) GPUs using vLLM~\citep{kwon2023efficient} for efficient batched inference. All models were evaluated using the same answer extraction and scoring pipeline. Each configuration was evaluated using a single inference run. The complete evaluation required approximately 300 GPU hours.

\paragraph{Decoding configuration.}
We use deterministic decoding (\textit{temperature = 0}, \textit{top-p = 1.0}) with a maximum generation length of 32{,}768 tokens. Generation terminates upon emitting the end-of-sequence (EOS) token or reaching the token limit. When the limit is reached without producing an answer, inference is rerun using the model's maximum supported context length.

\begin{figure*}[t]
    \centering
    \includegraphics[width=1\linewidth]{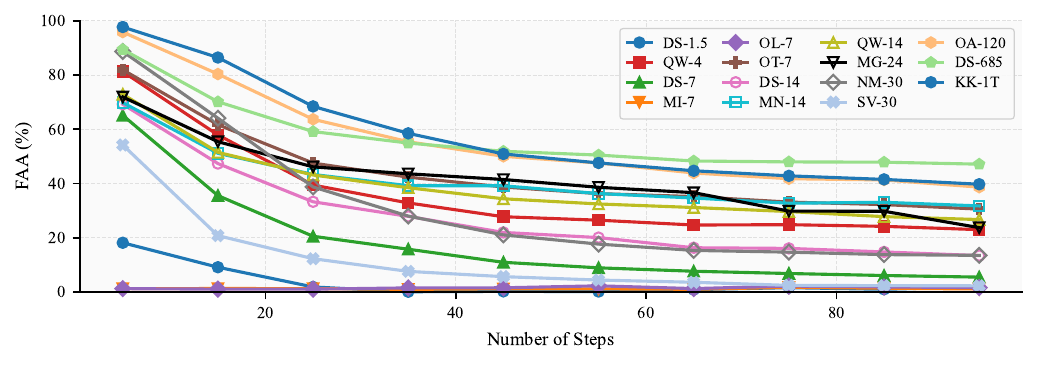}
    \caption{FAA of various language models as a function of procedure step count (5–95). Performance consistently declines with increasing steps for nearly all models, consistent with increasing difficulty in maintaining correct execution over longer arithmetic procedural sequences despite the simplicity of individual operations.}
    \label{fig:image4.1}
\end{figure*}

\begin{figure}[t]
    \centering
    \includegraphics[width=1\linewidth]{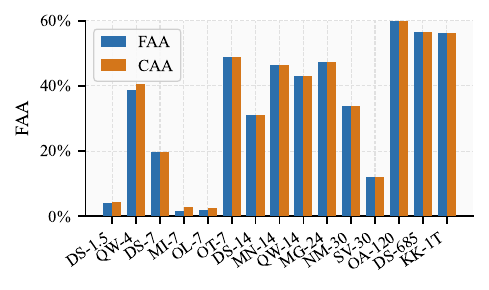}
    \caption{Comparison of FAA and CAA across models.}
    \label{fig:faa_vs_caa}
\end{figure}

\subsection{Evaluation Metrics}
\label{sec:evaluation_metrics}

We evaluate six complementary metrics: First-Answer Accuracy (FAA), Correct@Any (CAA), Parseable Answer Rate (PAR), Answer Position (AP), Step-Following Behavior (SFB), and Step-Level Verification (SLV). These metrics are designed to compare relative model behavior across controlled benchmark settings rather than establish an absolute measure of benchmark difficulty.

Let \(Y_n\) denote the reference output for example \(n\), and let \(\hat{Y}_n^{(i)}\) denote the \(i\)-th tagged answer in the model's output. We define \(r(x)=\mathrm{round}(x,3)\), and let \(\mathbf{1}[\cdot]\) denote the indicator function, which evaluates to 1 when the condition is true and 0 otherwise.

\paragraph{First-answer accuracy (FAA).}

FAA measures whether the first extracted answer is correct:
\[
\mathrm{FAA}
=
\frac{1}{N}
\sum_{n=1}^{N}
\mathbf{1}\left[
r(\hat{Y}_n^{(1)}) = r(Y_n)
\right]
\]

\paragraph{Correct@Any (CAA).}

CAA marks an example as correct if any extracted answer matches the reference output:
\[
\mathrm{CAA}
=
\frac{1}{N}
\sum_{n=1}^{N}
\mathbf{1}\left[
\exists i \;
r(\hat{Y}_n^{(i)})
=
r(Y_n)
\right].
\]

\paragraph{Parseable Answer Rate (PAR).}

PAR measures the percentage of examples containing at least one parseable \texttt{<answer>} span extracted using regular-expression-based parsing:
\[
\mathrm{PAR}
=
\frac{1}{N}
\sum_{n=1}^{N}
\mathbf{1}\left[
|\mathcal{A}_n| > 0
\right],
\]
where \(\mathcal{A}_n\) denotes the set of extracted answer spans for example \(n\).

\paragraph{Answer Position (AP).}

AP measures the normalized position of the first parseable \texttt{<answer>} span in the generated output. Let \(p_n\) denote the token index of the first parseable answer span and \(L_n\) denote the total output length. We compute:
\[
\mathrm{AP}
=
\frac{1}{N}
\sum_{n=1}^{N}
\frac{p_n}{L_n}\times 100.
\]

\paragraph{Step-Following Behavior (SFB).}

For outputs containing visible reasoning traces, we extract generated step counts by parsing occurrences of patterns of the form \texttt{Step k}. Let \(G_n\) denote the number of parsed generated steps and \(T_n\) denote the expected number of steps. We classify outputs as under-execution when \(G_n<T_n\), exact execution when \(G_n=T_n\), and over-execution when \(G_n>T_n\). We report the percentage of examples in each category. SFB reflects visible step markers only: it does not verify intermediate values (Section~\ref{sec:stepwise_error_analysis}).

\paragraph{Step-Level Verification (SLV).} SLV checks the intermediate values a model writes. A step is \emph{correct} if its value equals the specified operation applied to the model's own previous values (0.1\% tolerance); we report the \emph{correct prefix}, the fraction of steps before the first incorrect one, and the type of that error (Appendix~\ref{app:step_verification}).
\section{Results}
\label{sec:results}

\subsection{Performance Degradation with Arithmetic Procedural Complexity}


\label{sec:accuracy_complexity}

\subsubsection{Effect of Execution Horizon}
\label{sec:accuracy_steps}

We first analyze how FAA changes as the procedure length increases from 5 to 95. Across models and experimental settings, we observe a consistent decline in FAA as the number of steps increases (see Figure~\ref{fig:image4.1}). Averaged across models, the difference between 5-step and 95-step procedures is approximately 43 percentage points (pp). NM-30, DS-7, QW-4, KK-1T, OA-120, and DS-14 exhibit the largest degradation, each dropping by more than 55 percentage points. In contrast, MI-7 and OL-7 show minimal decline; however, both models also maintain very low FAA across all procedure lengths, leaving limited room for further degradation.

Figure~\ref{fig:faa_vs_caa} further compares FAA and CAA across models. Overall, the mean difference between the two metrics is only \(3.35 \times 10^{-4}\), indicating that models rarely recover after producing an incorrect first answer. However, a few smaller models exhibit noticeably larger FAA–CAA gaps, suggesting that the correct answer is occasionally generated later in the output despite an incorrect initial answer. This behavior is largely absent in larger reasoning models, whose generations tend to remain consistent once an initial answer is produced. These observations indicate that self-correction within a single generation is uncommon overall, but its limited occurrence appears to be model-dependent.

\subsubsection{Effect of Look-back Dependency}
\label{sec:accuracy_lookback}

\begin{figure}[t]
    \centering
    \includegraphics[width=1\linewidth]{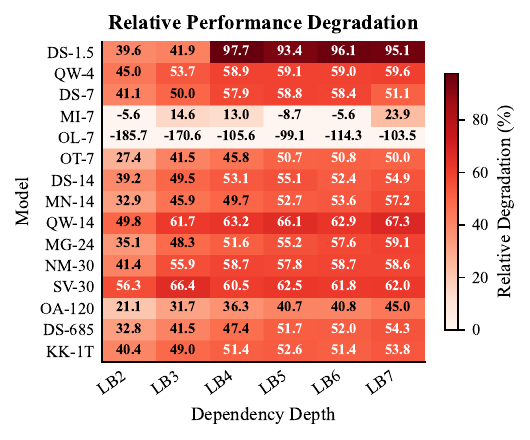}
    \caption{Relative FAA degradation (\%) on the mixed-operation benchmark only, measured with respect to each model's mixed-operation look-back 1 baseline. Each cell shows the percentage drop from the corresponding model’s single-step dependency baseline. Larger degradations indicate greater sensitivity to long-range intermediate-state dependencies.}
    \label{fig:image4.9}
\end{figure}

Figure~\ref{fig:image4.9} reports results using only the mixed-operation benchmark for all look-back settings, including the look-back 1 baseline, thereby isolating dependency depth from operation-type effects. Most models show progressively larger degradation as dependency depth increases, consistent with increasing difficulty in retrieving and maintaining non-local intermediate states. OA-120, DS-685, QW-14, MN-14, MG-24, and OT-7 achieve high FAA at shorter procedure lengths but degrade substantially with increasing complexity (see Figure~\ref{fig:lookback_heatmap} in Appendix~\ref{app:HeatMap}). MI-7 and OL-7 start near zero FAA, so their relative values are unstable; OL-7's are strongly negative because its FAA rises slightly from a near-zero baseline.

\subsection{Input and Task Factors Do Not Fully Explain the Degradation}
\label{sec:input_task_factors}

\begin{figure}[t]
    \centering
    \includegraphics[width=1\linewidth]{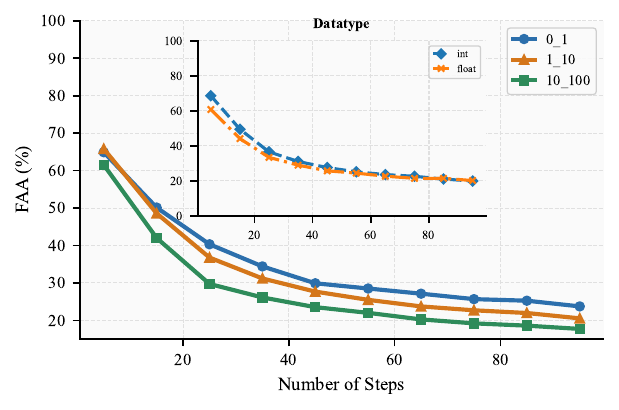}
    \caption{FAA across input ranges as a function of procedure length. The inset compares FAA trends for integer and floating-point input settings.}
    \label{fig:image4.65}
\end{figure}

\subsubsection{Effect of Input Range}
\label{sec:accuracy_ranges}

Figure~\ref{fig:image4.65} shows FAA across input ranges versus procedure length (averaged across models). FAA decreases with step count across all ranges. The $[0,1]$ range generally yields slightly higher FAA than the $[1,10]$ and $[10,100]$ ranges, improving by 4.14 pp on average.

Figure~\ref{fig:combined_range} (Appendix~\ref{app:Graphs}) shows that the median expected value does not strictly increase with procedure length. For the $[0,1]$ and $[10,100]$ ranges, correct predictions generally correspond to lower expected-value magnitudes, whereas the $[1,10]$ range shows a partially reversed trend at higher step counts. This suggests that performance differences across ranges cannot be explained solely by expected-output magnitude.

The model-wise heatmap in Figure~\ref{fig:image4.7} (Appendix~\ref{app:HeatMap}) reveals that smaller models (e.g., QW-4, DS-7) perform relatively better on the $[1,10]$ range, indicating that sensitivity to input range is model-dependent rather than uniform across architectures.




\subsubsection{Effect of Data Type}
\label{sec:accuracy_dtype}

The inset in Figure~\ref{fig:image4.65} compares FAA for integer and floating-point inputs. Both data types show a similar downward trend as the number of steps increases, with an average FAA difference of only 2.26 pp across models and procedure lengths. However, sensitivity varies across models. MN-14, SV-30, DS-14, and QW-14 perform better on integer inputs, with an average FAA gain of 6.53 pp, while OA-120, KK-1T, and OT-7 achieve higher FAA on floating-point inputs by an average of 6.34 pp. In contrast, DS-1.5, OL-7, NM-30, and MI-7 show minimal sensitivity to data type. Figure~\ref{fig:combined_dtype} in Appendix~\ref{app:Graphs} further shows that the median expected-output magnitude follows different trends across integer and floating-point settings for correct and incorrect predictions, suggesting that output magnitude alone does not consistently explain model failures.

\subsubsection{Effect of Task Type}
\label{sec:accuracy_tasks}

\begin{figure}[t]
    \centering
    \includegraphics[width=1\linewidth]{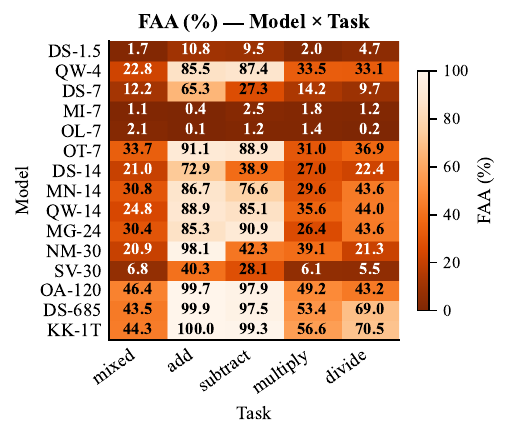}
    \caption{FAA heatmap across models and arithmetic task variants, including addition, subtraction, multiplication, division, and mixed-operation settings.}
    \label{fig:image4.105}
\end{figure}

Figure~\ref{fig:image4.105} shows model FAA across task variants. Averaged across models, addition and subtraction outperform multiplication and division by 34.74 pp, indicating strong operator sensitivity. This gap aligns with the output statistics in Figure~\ref{fig:task_types} (Appendix~\ref{app:Graphs}): multiplication and division exhibit substantially larger variance and heavier tails. For instance, multiplication outputs span over \(10^{190}\) in magnitude with unbounded variance, while division shows extreme spread due to near-zero denominators (Table~\ref{tab:task_output_stats}, Appendix~\ref{app:Graphs}). Such ill-conditioned distributions likely amplify intermediate errors and destabilize long-horizon execution. Additionally, the 10.09 pp FAA gap between addition and subtraction suggests models handle additive operations more reliably. Mixed-operation tasks further reduce average FAA by 22.06\% relative to single-operation settings, indicating added difficulty from operator switching.


\subsubsection{Expected-Output Magnitude Analysis}
\label{sec:expected_output_magnitude}

To examine whether numerical scale explains model failures, we compare the median expected output for correct and incorrect predictions across models. Figures~\ref{fig:group_small} to~\ref{fig:group_very_large} in Appendix~\ref{app:Graphs} show that the relationship between expected-output magnitude and correctness is not uniform across models. NM-30 and DS-7 achieve higher FAA on examples with larger expected values, while DS-1.5, DS-14, and OL-7 show only small differences between correct and incorrect predictions. Overall, expected-output magnitude may contribute to failures in some settings, particularly multiplication and division, but it does not fully explain the broader degradation with increasing arithmetic procedural complexity. Intermediate values show the same pattern (Appendix~\ref{app:intermediate_stats}): larger intermediate values lower FAA, but FAA still falls with length when all values stay below \(10^{2}\) (59.6\% to 33.6\%), and excluding examples with non-finite values changes FAA by at most 0.5 points.

\subsection{Generation-Level Failure Modes}
\label{sec:generation_failure_modes}

\subsubsection{Parseable Answer Rate}
\label{sec:non_null_outputs}




Table~\ref{tab:combined_degradation} and Figure~\ref{fig:image4.11} in Appendix~\ref{app:Others} show PAR as a function of procedure length. Averaged across models, PAR decreases by 24.63 pp from 5-step to 95-step procedures, indicating increasing difficulty in maintaining structured output generation over longer execution horizons. DS-7, SV-30, MG-24, NM-30, and DS-1.5 exhibit the largest PAR degradation, while OA-120, DS-685, KK-1T, OL-7, and QW-4 remain the most robust across increasing procedure lengths.

\subsubsection{Answer Position}
\label{sec:answer_position}



Figure~\ref{fig:image4.12} in Appendix~\ref{app:Others} shows the distribution of normalized answer positions across models. OA-120 frequently produces answers very early in the generation, often without explicitly generating intermediate execution steps (see Example~\ref{app:example_1} in Appendix~\ref{app:model_outputs}). In contrast, models such as QW-4, NM-30, and SV-30 also exhibit low normalized answer positions, but typically generate all intermediate steps before the first \texttt{<answer>} span. Their low normalized positions arise because the answer often appears within the first 100--200 tokens, followed by very long continuation traces. Thus, low answer position does not necessarily indicate shortcut reasoning or premature execution, but instead reflects differences in generation style and output structure.

\subsubsection{Step-wise Execution Breakdown}
\label{sec:stepwise_error_analysis}



\begin{table}[t]
\centering
\small
\setlength{\tabcolsep}{4pt}

\begin{tabular}{l c | c c c}
\toprule
\textbf{Model} & \textbf{PAR} $\downarrow$ & \textbf{Exact} $\downarrow$ & \textbf{Under} $\uparrow$ & \textbf{Over} \\
\midrule
OL-7   & -4.8 & 7.0  & -7.0  & 0.0 \\
QW-4   & 4.4  & 54.0 & -53.8 & -0.2 \\
KK-1T & 12.1 & 20.5 & -20.5 & 0.0 \\
DS-685 & 12.5 & 12.9 & -12.9 & 0.0 \\
OA-120 & 13.3 & 2.7  & -2.7  & 0.0 \\
MI-7   & 15.3 & 34.1 & -39.8 & +5.7 \\
OT-7   & 15.5 & 23.4 & -23.4 & 0.0 \\
QW-14  & 16.6 & 17.7 & -17.6 & -0.1 \\
MN-14  & 16.7 & 9.0  & -9.0  & 0.0 \\
DS-14  & 24.5 & 16.5 & -16.5 & 0.0 \\
DS-1.5 & 36.9 & 32.2 & -17.3 & -14.9 \\
NM-30  & 39.9 & 1.1  & -38.9 & +37.9 \\
MG-24  & 42.4 & 38.1 & -38.1 & 0.0 \\
SV-30  & 51.0 & 19.7 & -35.9 & +16.2 \\
DS-7   & 60.8 & 57.0 & -45.8 & -11.2 \\
\bottomrule
\end{tabular}

\caption{Change from 5-step to 95-step procedures (5-step minus 95-step, in pp); positive values indicate a decrease. Arrows show the typical direction of change with length. PAR: Parseable Answer Rate; Exact, Under, Over: step-following behavior.}
\label{tab:combined_degradation}

\end{table}

Table~\ref{tab:combined_degradation} shows the change in exact, under-, and over-execution rates between 5-step and 95-step procedures. As procedure length increases, exact execution consistently decreases while under-execution becomes more frequent. DS-7 and QW-4 exhibit the largest shifts, with exact execution dropping by 57.0 and 54.0 pp, respectively, whereas OA-120 and OL-7 remain comparatively stable. 
Inspecting 150 sampled outputs (10 per model), the SFB label matches the number of executed steps in 132 cases (88\%); errors arise when models restate the procedure without computing it (inflating exact execution, e.g., SV-30, MI-7) or compute steps without \texttt{Step~k} markers (e.g., NM-30). SFB thus reflects visible step generation: only 36.9\% of outputs labeled exact have all steps correct (Appendix~\ref{app:sfb_validation}).

\subsubsection{Step-Level Verification}
\label{sec:step_verification}

\begin{figure}[t]
    \centering
    \revcolor
    \includegraphics[width=1\linewidth]{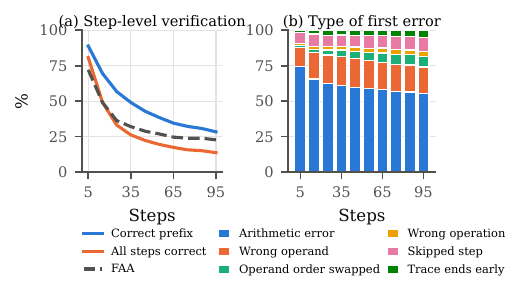}
    \caption{SLV for the 10 models that show intermediate steps in at least half of their outputs. (a) Correct prefix (\% of steps before the first error), share of examples with all steps correct, and FAA. (b) Type of the first error.}
    \label{fig:step_verification}
\end{figure}

Figure~\ref{fig:step_verification} applies SLV to the 10 models that show intermediate steps in at least half of their outputs (all models in Appendix~\ref{app:step_verification}). The correct prefix falls from 88.8\% of steps at 5 steps to 28.4\% at 95 (94.7\% to 36.1\% when matched directly against the reference executor; Appendix~\ref{app:step_verification}), and from 46.3\% at look-back 1 to 37.6\% at look-back 7, with most of the drop by look-back 4 (37.2\%). For seven of these models, 12--20\% of examples are answered correctly despite an incorrect intermediate step. Arithmetic slips are the most common first error, but state-tracking errors (wrong or swapped operands, skipped steps) rise from 22.5\% to 36.3\% of first errors as procedures grow longer. Some arithmetic errors are small approximations: at 1\% tolerance, the correct prefix of MG-24 rises from 51.7\% to 85.1\%. Formatting failures (no parseable answer) are counted separately: they make up 28.8\% of incorrect answers at 5 steps and 43.1\% at 95.

\section{Conclusion and Future Work}

In this work, we evaluated 15 instruction-tuned and reasoning models on a controlled benchmark for multi-step arithmetic procedural execution with varying procedure lengths and look-back dependencies. We find that performance consistently decreases as the arithmetic procedural complexity increases. Step-level verification confirms this: errors start early, and state-tracking errors grow with procedure length and dependency depth. Future work includes extending the benchmark to branching procedures, symbolic transformations, natural-language workflows, and tool-augmented agents to better understand faithful multi-step instruction execution beyond arithmetic settings.

\section{Limitations}
Our study is limited to synthetic arithmetic tasks and may not fully generalize to broader reasoning domains. We make no claim that these effects transfer directly to code execution or tool use. The benchmark does not include branching logic, semantic reasoning, external tool use, or interactive decision-making, and we do not evaluate commercial frontier models. SFB and SLV only see what models write, so they cannot assess outputs without visible steps (Appendix~\ref{app:step_verification}). While we observe a strong relationship between arithmetic procedural complexity and performance degradation, extending the analysis to more diverse structured reasoning tasks remains an important direction for future work.

\section{Ethical Considerations}
Our work evaluates language models on a fully synthetic benchmark for arithmetic procedural execution using automatically generated arithmetic tasks. Since the dataset is programmatically constructed, it does not contain personally identifiable information (PII), human annotations, or user-generated content. All evaluated models used in this study are publicly available open-source models and were accessed and evaluated in accordance with their respective licenses and intended research use. The benchmark is intended solely for research and evaluation purposes to study procedural reasoning and intermediate state tracking in language models. While the benchmark is designed to isolate procedural reasoning behavior in a controlled setting, it is limited to arithmetic execution and may not fully reflect broader real-world reasoning tasks. We hope this work contributes toward more reliable and transparent evaluation of long-horizon instruction-following behavior in language models.

\bibliography{custom, anthology-1, anthology-2}

\appendix
\onecolumn

\addtocontents{toc}{\protect\setcounter{tocdepth}{2}}
\startcontents[appendix]
\printcontents[appendix]{}{1}{\section*{Appendix Contents}}

\newpage

\section{Expected Output Computation}
\label{app:expected_output}

\begin{figure}[H]
\centering

\begin{minipage}{0.95\linewidth}
\begin{lstlisting}
def exec_algo(prompt, inputs):
    lines = prompt.strip().split("\n")

    vars_dict = {}

    vars_dict["S1"] = inputs[0]
    vars_dict["S2"] = inputs[1]

    for line in lines:
        line = line.strip()

        if "=" not in line:
            continue

        try:
            expr_part = line.split(":")[-1].strip()
            left, right = expr_part.split("=")

            left = left.strip()
            right = right.strip()

            first_var = right.split()[0].strip()
            task = right.split()[1].strip()
            second_var = right.split()[2].strip()
            value = eval(f"vars_dict['{first_var}'] {task} vars_dict['{second_var}']")

            vars_dict[left] = value

        except Exception as e:
            pass

    return [str(v) for v in vars_dict.values()]
\end{lstlisting}
\end{minipage}

\caption{Algorithm used to evaluate step-wise arithmetic procedures and compute deterministic reference outputs.}
\label{fig:exec_algo}

\end{figure}

\clearpage
\section{Prompt used during Inference}
\label{sec:prompt}

\begin{figure}[H]
\centering

\begin{minipage}{0.95\linewidth}
\begin{lstlisting}
You are a deterministic procedure executor.

You will be given:
1. A procedure (step-by-step instructions)
2. Input values

Your task is to simulate the procedure exactly as written.

Execution rules:
- Execute each step sequentially, without skipping or reordering.
- Maintain exact intermediate state.
- Do not reinterpret or optimize steps.

Behavior:
Show step-by-step execution.
Include intermediate variable values after each step.

Output rules:
- Show all intermediate steps clearly.
- Use the same variable names as in the procedure (S1, S2, etc.).
- That line must be in the format: <answer>FINAL_RESULT</answer>.
- Do not include any additional text before or after these tags.
- Do not use phrases like 'The final result is'.
- The output must contain only the tagged answer.
- Example (correct):
    <answer>42</answer>

Example (incorrect):
The final result is <answer>42</answer>
Answer: 42

Procedure:
function(x, y):
Let S1 = x
Let S2 = y
Step 1: S3 = S1 + S2
Step 2: S4 = S2 * S3
Step 3: S5 = S4 / S2
Step 4: S6 = S5 - S2
Step 5: S7 = S2 * S6
Final Step: Return S7

Inputs:
x = 5, y = 8
\end{lstlisting}
\end{minipage}

\caption{\rev{Exact} inference prompt used for arithmetic procedural execution experiments.}
\label{fig:prompt}

\end{figure}

\paragraph{Note.} The output rules ambiguously ask both for intermediate steps and for only the tagged answer. The intended format is the intermediate steps followed by a final \texttt{<answer>} line; scoring uses the first \texttt{<answer>} span wherever it appears.
\section{Results: Expected Output Analysis}
\label{app:Graphs}

This appendix provides a detailed analysis of how the magnitude of expected outputs relates to model performance across different settings. We report the median expected output as a function of the number of steps, separating correct and incorrect predictions to better understand whether numerical scale contributes to errors. These plots complement the main results by showing that while numerical magnitude can influence performance in certain cases (notably multiplication and division), it does not fully account for the observed degradation in accuracy with increasing arithmetic procedural complexity.

\begin{figure}[H]
    \centering
    \includegraphics[width=1\linewidth]{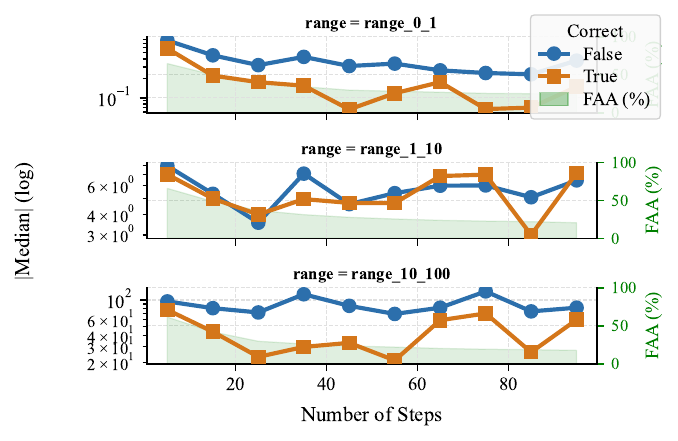}
    \caption{Median expected output across steps for different input ranges ([0,1], [1,10], [10,100]) separated by correct and incorrect model prediction.}
    \label{fig:combined_range}
\end{figure}

\begin{figure}[H]
    \centering
    \includegraphics[width=1\linewidth]{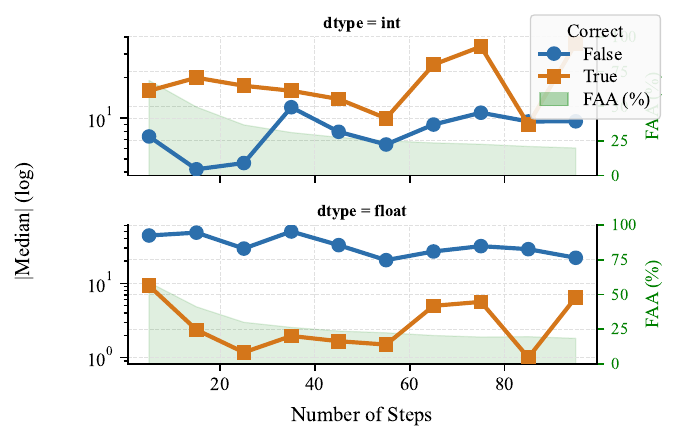}
    \caption{Median expected output across steps for integer and floating-point inputs, separated by correct and incorrect predictions.}
    \label{fig:combined_dtype}
\end{figure}

\clearpage

\begin{figure}[H]
    \centering
    \includegraphics[width=0.9\linewidth]{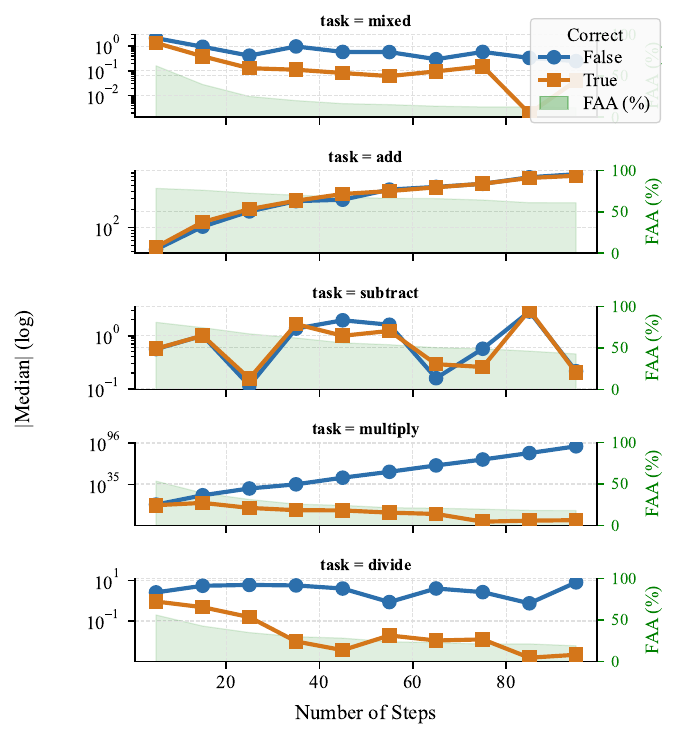}
    \caption{Median expected output across steps for different task types (addition, subtraction, multiplication, division, and mixed), grouped by correctness. Multiplication and division exhibit a larger separation between correct and incorrect predictions compared to other tasks.}
    \label{fig:task_types}
\end{figure}

\begin{table}[H]
\centering
\small

\begin{tabular}{lcccc}
\toprule
\textbf{Task} & \textbf{Median} & \textbf{Std. Dev.} & \textbf{Min} & \textbf{Max} \\
\midrule
Add      & \(3.45\times10^2\)  & \(1.91\times10^3\)  & \(1.02\times10^{-1}\) & \(9.56\times10^3\) \\
Subtract & \(-2.89\times10^{-1}\) & \(2.48\times10^2\)  & \(-1.82\times10^3\) & \(1.36\times10^3\) \\
Multiply & \(6.25\times10^{31}\) & \(\infty\) & \(0\) & \(1.43\times10^{190}\) \\
Divide   & \(8.75\times10^{-1}\) & \(2.71\times10^{33}\) & \(1.82\times10^{-44}\) & \(1.40\times10^{35}\) \\
Mixed    & \(9.97\times10^{-1}\) & \(7.80\times10^{29}\) & \(-3.61\times10^{29}\) & \(3.98\times10^{31}\) \\
\bottomrule
\end{tabular}

\caption{Expected-output statistics across task variants. Multiplication and division exhibit substantially larger variance and heavier-tailed output distributions compared to addition and subtraction.}
\label{tab:task_output_stats}

\end{table}

\begin{figure}[H]
    \centering

    \begin{subfigure}{0.9\linewidth}
        \centering
        \includegraphics[width=\linewidth]{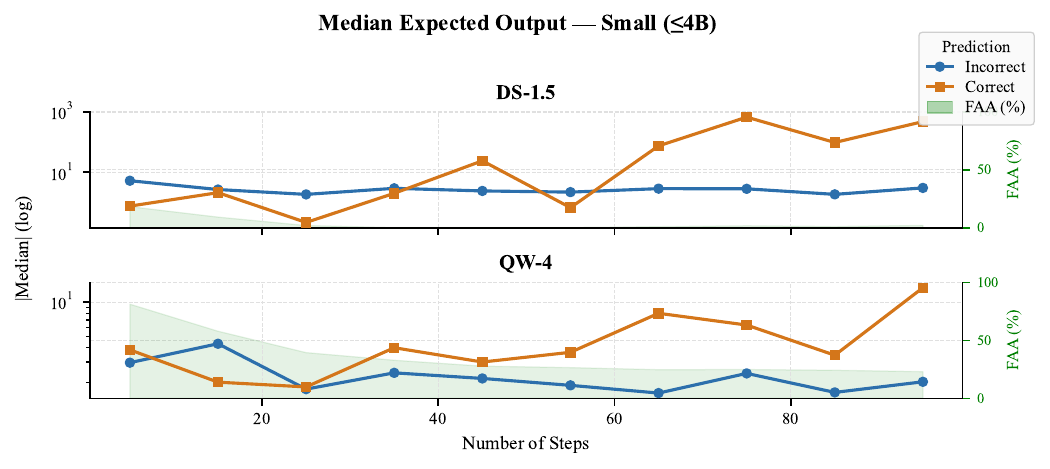}
        \caption{Small models ($\leq$4B)}
        \label{fig:small}
    \end{subfigure}

    \vspace{0.5em}

    \begin{subfigure}{0.9\linewidth}
        \centering
        \includegraphics[width=\linewidth]{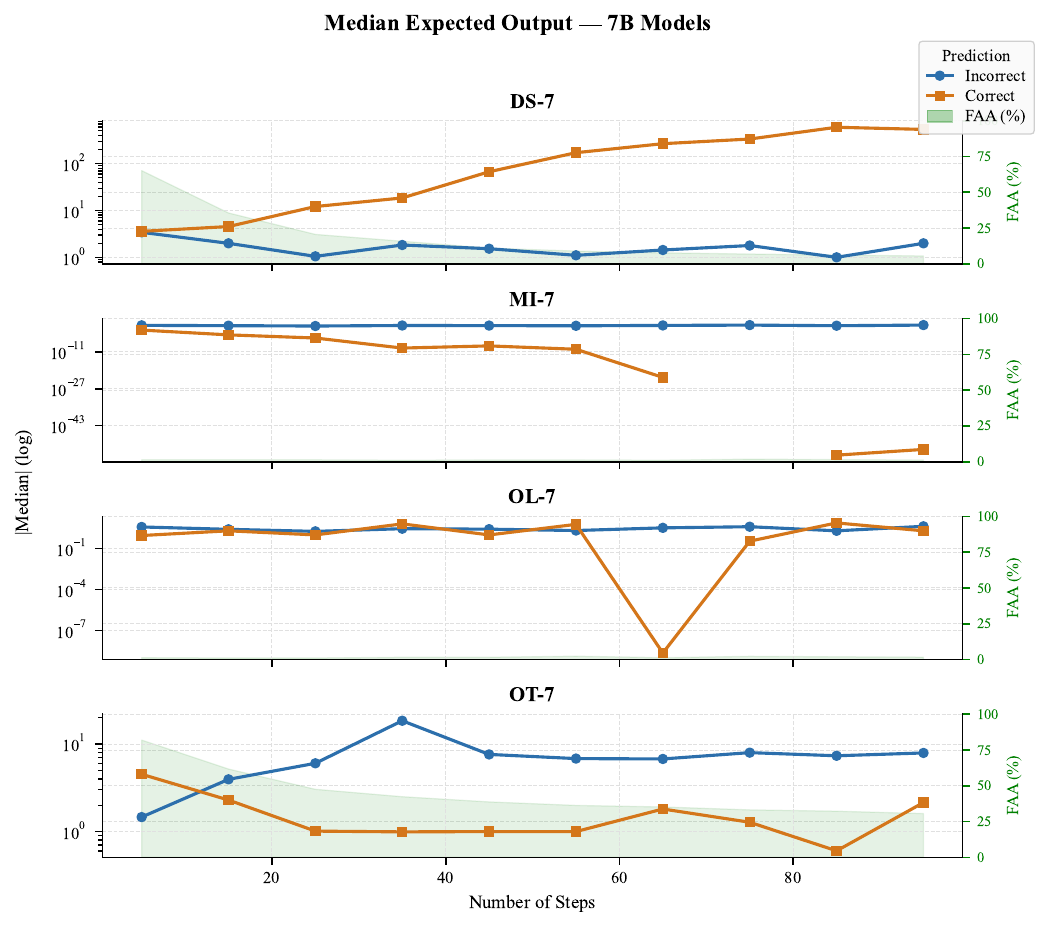}
        \caption{Mid-scale models (7B)}
        \label{fig:7b}
    \end{subfigure}

    \caption{Median expected output across steps for small ($\leq$4B) and mid-scale (7B) models. Trends vary across models, with no consistent relationship between output magnitude and correctness.}
    \label{fig:group_small}
\end{figure}

\begin{figure}[H]
    \centering

    \begin{subfigure}{0.9\linewidth}
        \centering
        \includegraphics[width=\linewidth]{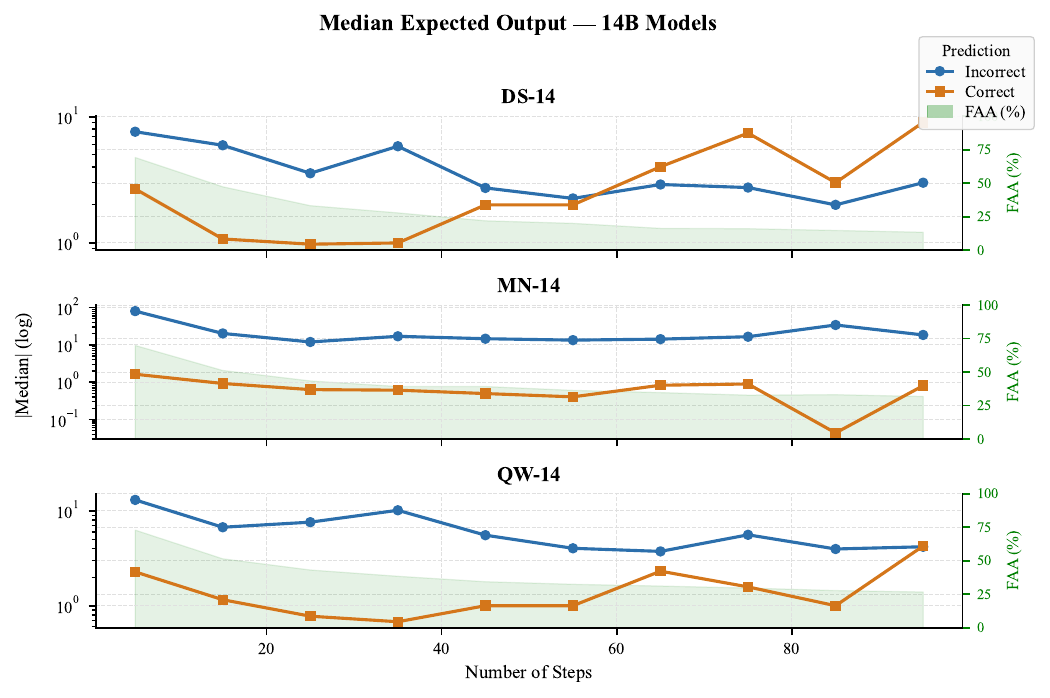}
        \caption{Large models (14B)}
        \label{fig:14b}
    \end{subfigure}

    \vspace{0.5em}

    \begin{subfigure}{0.9\linewidth}
        \centering
        \includegraphics[width=\linewidth]{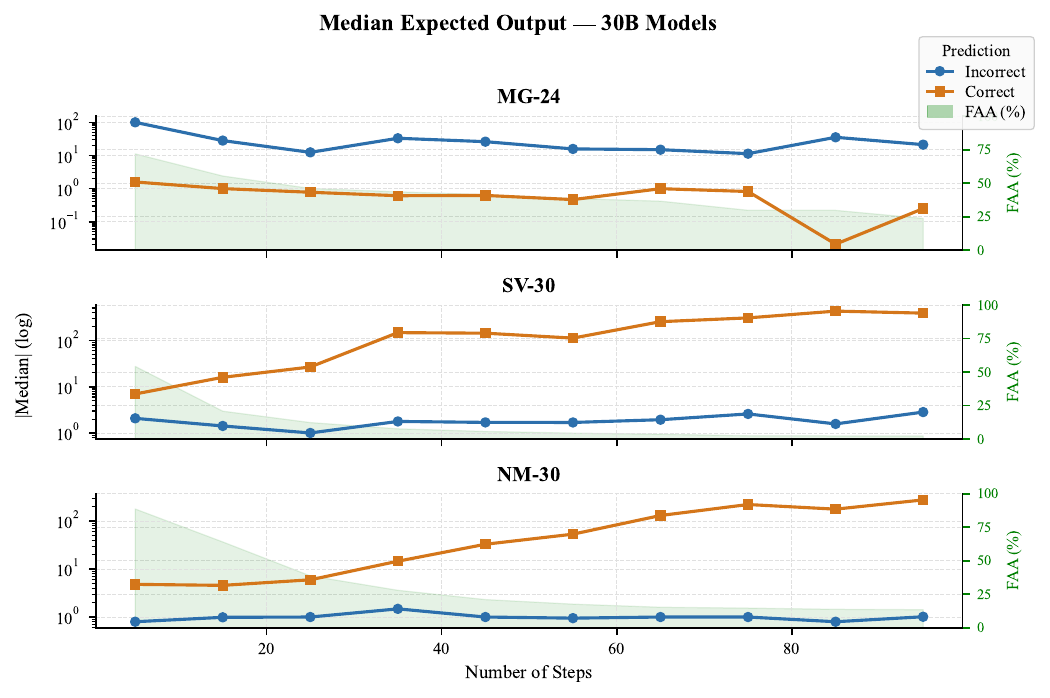}
        \caption{Very large models (24B--30B)}
        \label{fig:30b}
    \end{subfigure}

    \caption{Median expected output across steps for large (14B) and very large (24B--30B) models. While some models show sensitivity to output magnitude, the overall trend indicates that numerical scale alone does not explain performance degradation.}
    \label{fig:group_large}
\end{figure}

\begin{figure}[H]
    \centering

    \begin{subfigure}{0.9\linewidth}
        \centering
        \includegraphics[width=\linewidth]{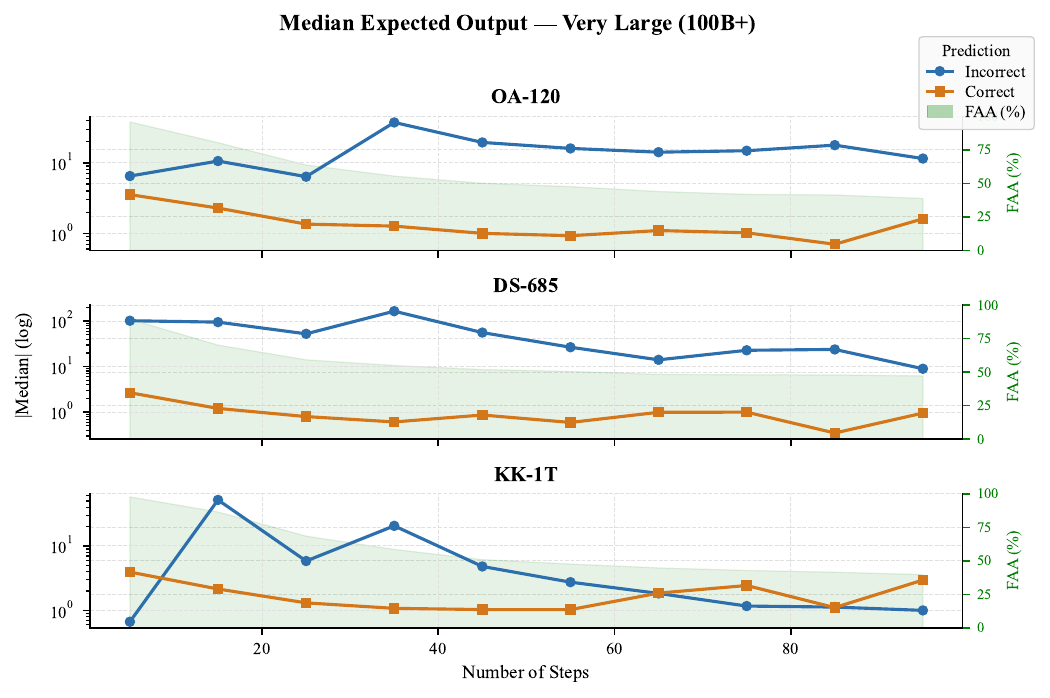}
        \caption{Frontier-scale models ($\geq$100B)}
        \label{fig:100b}
    \end{subfigure}

    \caption{Median expected output across steps for frontier-scale models ($\geq$100B). While some models show sensitivity to output magnitude, the overall trend indicates that numerical scale alone does not explain performance degradation.}
    \label{fig:group_very_large}
\end{figure}

\section{Results: Heatmaps}
\label{app:HeatMap}

\begin{figure}[H]
    \centering
    \includegraphics[width=0.7\linewidth]{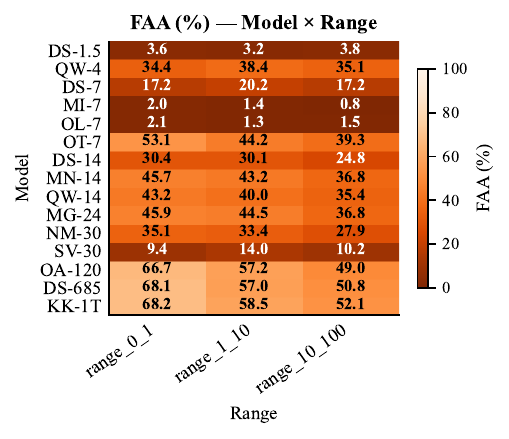}
    \caption{FAA heatmap across models and input ranges. Performance varies across models, with no uniform sensitivity to range; some models achieve higher FAA in intermediate ranges ([1,10]) than in smaller or larger ranges.}
    \label{fig:image4.7}
\end{figure}

\begin{figure}[H]
    \centering
    \includegraphics[width=0.7\linewidth]{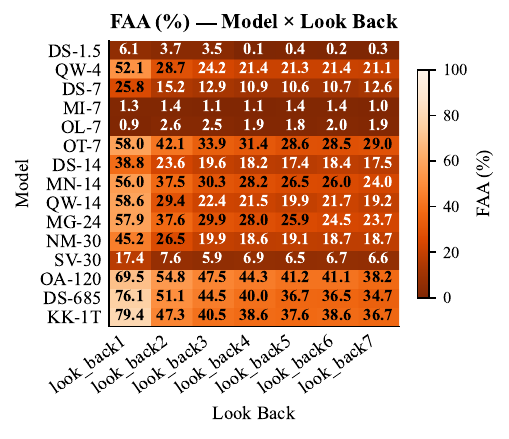}
    \caption{FAA heatmap across models and look-back dependency.}
    \label{fig:lookback_heatmap}
\end{figure}
\newpage
\section{Results: Generation-Level Failure Modes}
\label{app:Others}

\begin{figure}[H]
    \centering
    \includegraphics[width=1\linewidth]{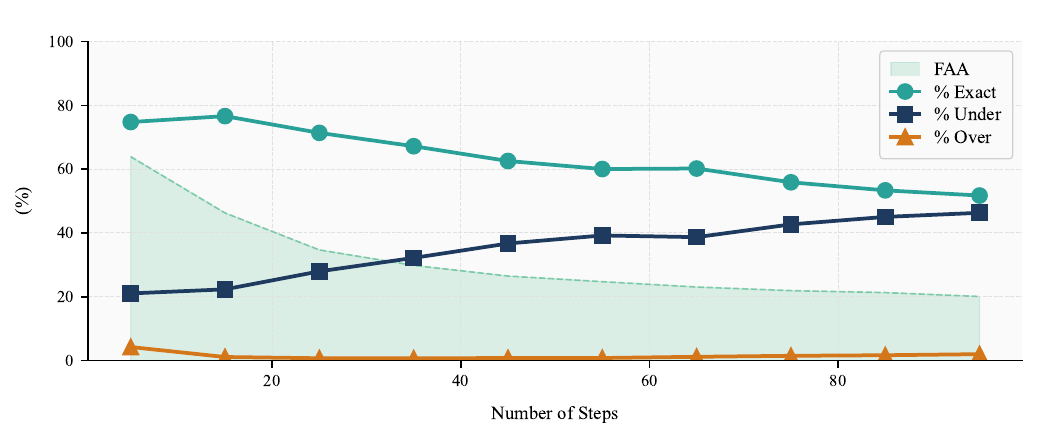}
    \caption{Step-following behavior and FAA across procedure lengths. As FAA (shaded) declines, exact executions decrease and under-executed generations rise, while over-execution remains rare.}
    \label{fig:image4.13}
\end{figure}

\begin{figure}[H]
    \centering
    \includegraphics[width=1\linewidth]{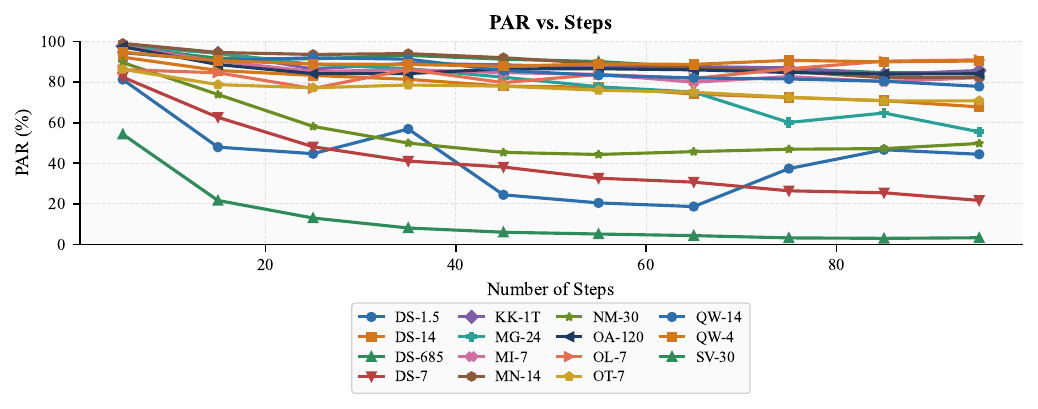}
    \caption{Parseable Answer Rate (PAR) across increasing step counts. While many models maintain high PAR, several exhibit declining rates with longer procedures, indicating a growing tendency to fail in producing a valid structured answer as complexity increases.}
    \label{fig:image4.11}
\end{figure}

\begin{figure}[H]
    \centering
    \includegraphics[width=1\linewidth]{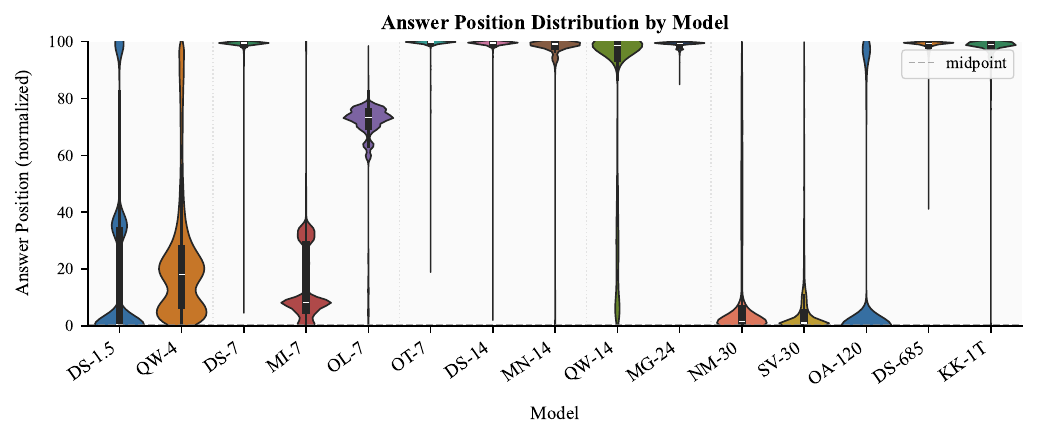}
    \caption{Distribution of the normalized position of the first generated answer across models. Models vary in when they commit to an answer, with some producing early answers and others delaying until later in the output, reflecting different generation strategies rather than consistent differences in correctness.}
    \label{fig:image4.12}
\end{figure}

\clearpage
\section{Step-Level Verification: Method and Additional Results}
\label{app:step_verification}

\subsection{Method}
\paragraph{Data.} SLV uses the stored generation of every evaluated example (Table~\ref{tab:coverage}). Generations of DS-685 and KK-1T were stored only for 13 and 2 look-back-1 configurations, so their results are not comparable to those of the other models (Table~\ref{tab:c2_common_subset} compares all models on shared configurations). Reference values are those each output was scored against. For 4{,}000 QW-4 examples whose procedure text was not stored, each step's operation and operands were reconstructed from the reference values.

\paragraph{Extracting intermediate values.} For each variable $S_t$ ($t \geq 3$) we take the \emph{first} value the model commits to, e.g., \texttt{S3 = 6.049 + 7.183 = 13.232}, \texttt{S3 $\approx$ 13.232}, \texttt{S3 is 13.232}, a table row, or a step written in prose (\texttt{Step 1: S3 = S1 + S2. That's 2.679 + 9.704, which equals 12.383}). Restating the procedure (\texttt{Step 1: S3 = S1 + S2}) or an unfinished expression does not count as a value.

\paragraph{Correct steps and error types.} For step $S_t = S_a \circ S_b$ with model values $m$, the step is \emph{correct} if $m_t$ equals $m_a \circ m_b$ within 0.1\% (or within one unit of the last digit written), so rounding carried over from earlier steps is not counted as an error. The \emph{correct prefix} is the fraction of steps before the first incorrect one. That step is classified as \emph{skipped} ($m_t$ missing but later values present), \emph{trace ends} (no further values), \emph{operand order swapped}, \emph{wrong operation}, \emph{wrong operand} (result of other available variables), or otherwise \emph{arithmetic error}. Evaluation stops at the first non-finite reference value; outputs without any intermediate value are excluded and reported as ``visible states''.

\begin{table}[H]
\small\centering\revcolor
\caption{Stored generations used for the intermediate-state analysis.}
\label{tab:coverage}
\begin{tabular}{lrrr}
\toprule
 & Examples & Configurations & Procedure reconstructed \\
\midrule
DS-1.5 & 55000 & 55 & 0 \\
QW-4 & 55000 & 55 & 4000 \\
OL-7 & 55000 & 55 & 0 \\
MI-7 & 55000 & 55 & 0 \\
DS-7 & 54500 & 55 & 0 \\
OT-7 & 54500 & 55 & 0 \\
MN-14 & 55000 & 55 & 0 \\
DS-14 & 55000 & 55 & 0 \\
QW-14 & 55000 & 55 & 0 \\
MG-24 & 55000 & 55 & 0 \\
SV-30 & 55000 & 55 & 0 \\
NM-30 & 55000 & 55 & 0 \\
OA-120 & 55000 & 55 & 0 \\
DS-685 & 12768 & 13 & 0 \\
KK-1T & 777 & 2 & 0 \\
\bottomrule
\end{tabular}
\end{table}

\subsection{Additional Results}
\begin{figure}[H]
    \centering
    \revcolor
    \begin{minipage}{0.48\linewidth}
        \centering
        \includegraphics[width=\linewidth]{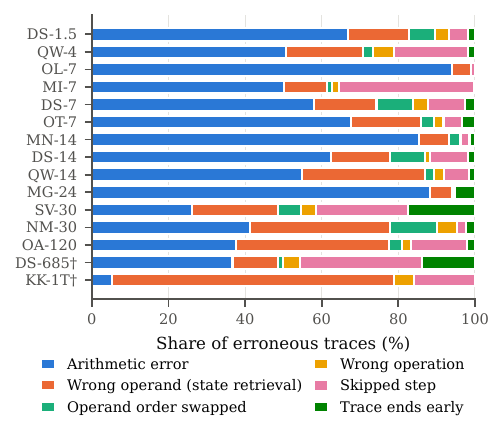}
    \end{minipage}\hfill
    \begin{minipage}{0.48\linewidth}
        \centering
        \includegraphics[width=\linewidth]{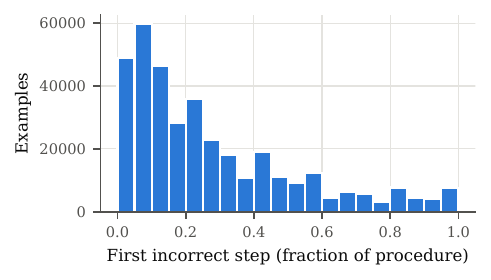}
    \end{minipage}
    \caption{Left: type of the first error per model. Right: position of the first error relative to procedure length, pooled over the models in Figure~\ref{fig:step_verification}.}
    \label{fig:c8_error_types}
\end{figure}

Matching values directly against the reference (1\% tolerance) gives the same trend as the correct prefix (94.7\% to 36.1\% of steps from 5 to 95 steps). For most models, the first error lies in the first 10--40\% of the procedure (Figure~\ref{fig:c8_error_types}). Wrong-operand errors increase with look-back depth (14.8\% of first errors at look-back 1 vs.\ 22--26\% at look-back 2--7; Table~\ref{tab:c8_error_types_by_lookback}). Part of the arithmetic errors are small approximations: with a 1\% instead of 0.1\% tolerance, the correct prefix of MG-24 rises from 51.7\% to 85.1\% (Table~\ref{tab:c2_tolerance_sensitivity}). OA-120, MI-7, and OL-7 show intermediate values in only 15.5\%, 10.5\%, and 1.5\% of outputs, so their statistics cover a small subset.

\begin{table}[H]
\small\centering\revcolor
\caption{Intermediate-state verification against the reference executor. Prefix metrics are computed over examples with at least one parseable intermediate state. $^\dagger$Traces available for a subset only.}
\label{tab:c2_intermediate_state_by_model}
\resizebox{\linewidth}{!}{\begin{tabular}{lrlllllll}
\toprule
 & N & FAA (\%) & Visible states (\%) & State coverage (\%) & Ref.-matching prefix (\%) & Correct prefix (\%) & All steps correct (\%) & Median first error (rel.) \\
\midrule
DS-1.5 & 55000 & 3.9 & 57.5 & 77.7 & 37.4 & 26.8 & 12.8 & 0.1 \\
QW-4 & 55000 & 38.6 & 100.0 & 72.9 & 52.4 & 46.8 & 30.9 & 0.2 \\
OL-7 & 55000 & 2.0 & 1.5 & 98.7 & 88.7 & 57.6 & 36.0 & 0.4 \\
MI-7 & 55000 & 1.5 & 10.5 & 71.8 & 31.1 & 25.9 & 14.2 & 0.2 \\
DS-7 & 54500 & 19.7 & 97.5 & 81.8 & 46.6 & 35.9 & 19.2 & 0.1 \\
OT-7 & 54500 & 49.0 & 100.0 & 92.1 & 75.9 & 67.8 & 47.1 & 0.4 \\
MN-14 & 55000 & 46.4 & 98.3 & 95.9 & 72.1 & 52.0 & 33.2 & 0.2 \\
DS-14 & 55000 & 31.0 & 99.7 & 88.8 & 54.9 & 44.1 & 25.3 & 0.2 \\
QW-14 & 55000 & 42.9 & 100.0 & 92.1 & 65.1 & 55.9 & 36.7 & 0.2 \\
MG-24 & 55000 & 47.2 & 100.0 & 93.7 & 75.6 & 51.7 & 33.5 & 0.2 \\
SV-30 & 55000 & 12.1 & 92.5 & 56.6 & 42.7 & 42.2 & 26.0 & 0.2 \\
NM-30 & 55000 & 33.8 & 99.9 & 81.2 & 43.1 & 42.3 & 23.8 & 0.2 \\
OA-120 & 55000 & 60.0 & 15.5 & 95.5 & 88.4 & 86.4 & 69.9 & 0.6 \\
DS-685$^\dagger$ & 12768 & 81.5 & 100.0 & 99.1 & 97.6 & 97.4 & 95.1 & 0.4 \\
KK-1T$^\dagger$ & 777 & 98.7 & 99.5 & 100.0 & 99.2 & 99.1 & 97.5 & 0.8 \\
\bottomrule
\end{tabular}}
\end{table}

\begin{table}[H]
\small\centering\revcolor
\caption{Relation between correct step execution and final-answer correctness (examples with visible states). Steps: all steps correct; FAA: first answer correct.}
\label{tab:c2_faithfulness_vs_correctness}
\begin{tabular}{llll}
\toprule
 & P(Steps $|$ FAA) (\%) & P(FAA $|$ Steps) (\%) & Correct answer, incorrect step (\%) \\
\midrule
DS-1.5 & 68.0 & 34.5 & 2.1 \\
QW-4 & 67.4 & 84.4 & 12.6 \\
OL-7 & 66.7 & 1.4 & 0.2 \\
MI-7 & 23.8 & 3.5 & 1.6 \\
DS-7 & 65.1 & 67.2 & 6.9 \\
OT-7 & 71.3 & 74.1 & 14.1 \\
MN-14 & 58.7 & 83.3 & 19.5 \\
DS-14 & 59.8 & 73.2 & 12.5 \\
QW-14 & 66.9 & 78.1 & 14.2 \\
MG-24 & 61.1 & 86.2 & 18.4 \\
SV-30 & 89.8 & 45.1 & 1.3 \\
NM-30 & 62.8 & 89.1 & 12.6 \\
OA-120 & 82.2 & 76.0 & 11.5 \\
DS-685 & 97.7 & 83.8 & 1.9 \\
KK-1T & 97.6 & 98.8 & 2.3 \\
\bottomrule
\end{tabular}
\end{table}

\begin{figure}[H]
    \centering
    \revcolor
    \includegraphics[width=0.95\linewidth]{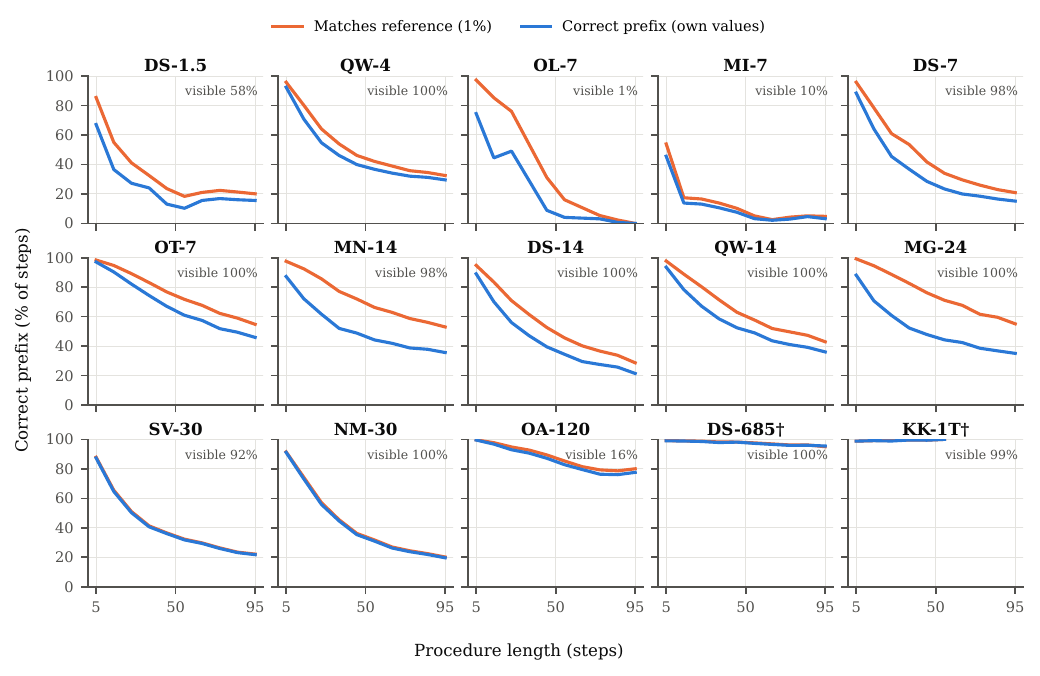}
    \caption{Correct prefix and reference-matching prefix by procedure length for each model (outputs with visible intermediate values). $\dagger$: subset of configurations only.}
    \label{fig:c2_prefix_vs_steps}
\end{figure}

\begin{table}[H]
\small\centering\revcolor
\caption{Category of the first incorrect step (\% of examples with visible intermediate states).}
\label{tab:c8_error_types_by_model}
\resizebox{\linewidth}{!}{\begin{tabular}{lllllllll}
\toprule
 & Visible states (\%) & All steps correct & Arithmetic error & Wrong operand (state retrieval) & Operand order swapped & Wrong operation & Skipped step & Trace ends early \\
\midrule
DS-1.5 & 57.5 & 12.8 & 58.4 & 13.9 & 5.7 & 3.2 & 4.3 & 1.7 \\
QW-4 & 100.0 & 30.9 & 35.0 & 13.9 & 1.8 & 3.7 & 13.4 & 1.2 \\
OL-7 & 1.5 & 36.0 & 60.2 & 3.0 & 0.0 & 0.1 & 0.6 & 0.0 \\
MI-7 & 10.5 & 14.2 & 43.0 & 9.6 & 1.2 & 1.6 & 30.3 & 0.1 \\
DS-7 & 97.5 & 19.2 & 46.9 & 13.1 & 7.6 & 3.2 & 7.8 & 2.1 \\
OT-7 & 100.0 & 47.1 & 35.7 & 9.7 & 1.8 & 1.3 & 2.6 & 1.8 \\
MN-14 & 98.3 & 33.2 & 57.0 & 5.2 & 1.9 & 0.2 & 1.5 & 0.9 \\
DS-14 & 99.7 & 25.3 & 46.6 & 11.4 & 6.8 & 0.9 & 7.4 & 1.4 \\
QW-14 & 100.0 & 36.7 & 34.7 & 20.2 & 1.5 & 1.7 & 4.1 & 1.0 \\
MG-24 & 100.0 & 33.5 & 58.7 & 3.9 & 0.2 & 0.1 & 0.2 & 3.4 \\
SV-30 & 92.5 & 26.0 & 19.4 & 16.5 & 4.5 & 2.9 & 17.8 & 12.9 \\
NM-30 & 99.9 & 23.8 & 31.5 & 27.9 & 9.2 & 4.0 & 1.8 & 1.8 \\
OA-120 & 15.5 & 69.9 & 11.3 & 12.0 & 1.1 & 0.7 & 4.4 & 0.6 \\
DS-685$^\dagger$ & 100.0 & 95.1 & 1.8 & 0.6 & 0.1 & 0.2 & 1.6 & 0.7 \\
KK-1T$^\dagger$ & 99.5 & 97.5 & 0.1 & 1.8 & 0.0 & 0.1 & 0.4 & 0.0 \\
\bottomrule
\end{tabular}}
\end{table}

\begin{table}[H]
\small\centering\revcolor
\caption{Intermediate-state metrics by look-back depth (mixed operations).}
\label{tab:c2_by_lookback_pooled}
\resizebox{\linewidth}{!}{\begin{tabular}{lllllr}
\toprule
Look-back & Correct prefix (\%) & Ref.-matching prefix (\%) & All steps correct (\%) & FAA (\%) & Examples \\
\midrule
1 & 46.3 & 60.7 & 27.2 & 32.4 & 46598 \\
2 & 41.7 & 54.4 & 20.8 & 33.4 & 46197 \\
3 & 39.7 & 51.9 & 18.3 & 28.1 & 46028 \\
4 & 37.2 & 49.1 & 16.0 & 24.2 & 47841 \\
5 & 37.8 & 49.4 & 15.9 & 23.0 & 47791 \\
6 & 36.7 & 49.2 & 15.1 & 22.2 & 47796 \\
7 & 37.6 & 50.1 & 15.7 & 22.1 & 46722 \\
\bottomrule
\end{tabular}}
\end{table}

\begin{table}[H]
\small\centering\revcolor
\caption{Error-type composition by look-back depth (mixed operations, erroneous traces).}
\label{tab:c8_error_types_by_lookback}
\resizebox{\linewidth}{!}{\begin{tabular}{lllllll}
\toprule
Look-back & Arithmetic error & Wrong operand (state retrieval) & Operand order swapped & Wrong operation & Skipped step & Trace ends early \\
\midrule
1 & 64.7 & 14.8 & 3.1 & 6.2 & 7.6 & 3.6 \\
2 & 60.5 & 22.5 & 1.2 & 3.4 & 8.6 & 3.7 \\
3 & 59.1 & 25.5 & 0.9 & 2.9 & 8.0 & 3.6 \\
4 & 61.1 & 24.3 & 0.9 & 3.0 & 7.3 & 3.5 \\
5 & 61.2 & 24.4 & 0.9 & 2.6 & 7.5 & 3.4 \\
6 & 63.1 & 23.3 & 0.9 & 2.4 & 7.1 & 3.3 \\
7 & 63.2 & 23.6 & 0.9 & 2.2 & 6.7 & 3.4 \\
\bottomrule
\end{tabular}}
\end{table}

\begin{table}[H]
\small\centering\revcolor
\caption{Intermediate-state metrics on the configurations available for all 15 models (range 0--1, float, look-back 1, add and mixed).}
\label{tab:c2_common_subset}
\begin{tabular}{lrlll}
\toprule
 & N & Correct prefix (\%) & All steps correct (\%) & FAA (\%) \\
\midrule
DS-1.5 & 888 & 49.8 & 36.3 & 15.1 \\
QW-4 & 1998 & 60.8 & 52.0 & 27.7 \\
OL-7 & 83 & 81.8 & 69.9 & 0.0 \\
MI-7 & 214 & 13.1 & 6.1 & 0.0 \\
DS-7 & 1901 & 42.7 & 31.2 & 36.0 \\
OT-7 & 1997 & 77.8 & 66.8 & 79.8 \\
MN-14 & 1994 & 64.9 & 55.0 & 71.3 \\
DS-14 & 1985 & 54.1 & 43.1 & 57.9 \\
QW-14 & 1999 & 73.2 & 63.3 & 74.6 \\
MG-24 & 2000 & 66.2 & 52.5 & 71.8 \\
SV-30 & 1885 & 54.6 & 44.3 & 21.2 \\
NM-30 & 1998 & 59.1 & 47.2 & 62.6 \\
OA-120 & 272 & 96.8 & 94.1 & 93.0 \\
DS-685 & 2000 & 99.6 & 99.3 & 96.8 \\
KK-1T & 773 & 99.1 & 97.5 & 98.7 \\
\bottomrule
\end{tabular}
\end{table}

\begin{table}[H]
\small\centering\revcolor
\caption{Sensitivity of prefix metrics (mean \% of steps) to tolerance and to using the last instead of the first value per variable.}
\label{tab:c2_tolerance_sensitivity}
\resizebox{\linewidth}{!}{\begin{tabular}{lllllll}
\toprule
 & Ref. 0.1\% & Ref. 1\% & Ref. 5\% & Ref. 1\% (last value) & Correct prefix (0.1\%/ulp) & Correct prefix (1\%) \\
\midrule
DS-1.5 & 25.9 & 37.4 & 41.1 & 35.8 & 26.8 & 39.1 \\
QW-4 & 45.7 & 52.4 & 54.0 & 48.2 & 46.8 & 53.3 \\
OL-7 & 55.0 & 88.7 & 92.4 & 88.9 & 57.6 & 91.5 \\
MI-7 & 25.3 & 31.1 & 35.2 & 27.2 & 25.9 & 31.7 \\
DS-7 & 33.5 & 46.6 & 49.5 & 42.1 & 35.9 & 48.8 \\
OT-7 & 62.8 & 75.9 & 78.5 & 77.8 & 67.8 & 78.1 \\
MN-14 & 48.3 & 72.1 & 77.9 & 72.1 & 52.0 & 78.7 \\
DS-14 & 39.7 & 54.9 & 58.7 & 54.8 & 44.1 & 57.8 \\
QW-14 & 51.1 & 65.1 & 67.4 & 65.0 & 55.9 & 67.4 \\
MG-24 & 48.7 & 75.6 & 83.0 & 75.2 & 51.7 & 85.1 \\
SV-30 & 42.0 & 42.7 & 43.1 & 46.3 & 42.2 & 42.7 \\
NM-30 & 42.2 & 43.1 & 44.1 & 54.0 & 42.3 & 43.0 \\
OA-120 & 80.5 & 88.4 & 89.6 & 88.2 & 86.4 & 89.7 \\
DS-685 & 94.0 & 97.6 & 97.8 & 97.7 & 97.4 & 97.8 \\
KK-1T & 99.0 & 99.2 & 99.3 & 99.2 & 99.1 & 99.2 \\
\bottomrule
\end{tabular}}
\end{table}

\subsection{Validation of Step-Following Behavior}
\label{app:sfb_validation}
We inspected 150 outputs (10 per model, random) and counted the steps each actually computes. The SFB label matches this count in 132 cases (88\%). Of the 18 mismatches, 14 are overcounts: the output restates the procedure as \texttt{Step~k} lines but computes only some of the steps (SV-30: 5, QW-4: 4, NM-30: 2, DS-7, DS-1.5, MI-7: 1 each). Four are undercounts: steps computed without \texttt{Step~k} markers (DS-14, NM-30, QW-4, OA-120). Table~\ref{tab:sfb_validation} shows the same at scale: 8.8\% of outputs labeled exact contain values for fewer than half of the steps (up to 49.1\% for MI-7 and 42.5\% for SV-30), and 6.7\% of outputs labeled under-executed contain all values (up to 28.2\% for NM-30). Only 36.9\% of outputs labeled exact have all steps correct.

\begin{table}[H]
\small\centering
\revcolor
\caption{SFB labels compared with intermediate values extracted for SLV (all models).}
\label{tab:sfb_validation}
\begin{tabular}{lrrrr}
\toprule
SFB label & Outputs & All values present (\%) & $<$ half of values (\%) & All steps correct (\%) \\
\midrule
Exact execution & 433{,}818 & 82.3 & 8.8  & 36.9 \\
Over-execution  & 11{,}835  & 81.6 & 9.5  & 30.5 \\
Under-execution & 281{,}892 & 6.7  & 82.5 & 3.5 \\
\bottomrule
\end{tabular}
\end{table}

\clearpage
\section{Intermediate-Value Statistics}
\label{app:intermediate_stats}
\begin{figure}[H]
    \centering
    \revcolor
    \includegraphics[width=0.5\linewidth]{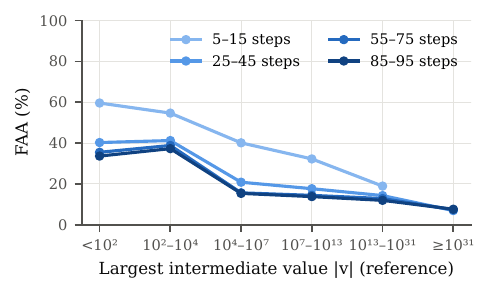}
    \caption{FAA by magnitude of the largest intermediate value of the reference execution, for four procedure-length groups.}
    \label{fig:c5_magnitude}
\end{figure}

For each example we compute, over the intermediate values of the reference execution, $\lfloor \log_{10} |v| \rfloor$ of the largest and of the smallest non-zero value, and whether any value is non-finite. Figure~\ref{fig:c5_magnitude} and Table~\ref{tab:c5_faa_by_magnitude_and_length} relate the largest intermediate value to FAA within procedure-length groups (13 models with complete generations; bins with fewer than 200 examples omitted from the figure). Excluding examples with non-finite values changes FAA by at most 0.5 points (Table~\ref{tab:c5_faa_nonfinite_sensitivity}).

\begin{table}[H]
\small\centering\revcolor
\caption{FAA (\%) by magnitude of the largest reference intermediate value, within procedure-length groups (13 full-trace models).}
\label{tab:c5_faa_by_magnitude_and_length}
\begin{tabular}{lllll}
\toprule
Largest $|v|$ & 5–15 & 25–45 & 55–75 & 85–95 \\
\midrule
$<$10$^{2}$ & 59.6 & 40.3 & 35.5 & 33.6 \\
10$^{2}$–10$^{4}$ & 54.7 & 41.2 & 38.8 & 37.3 \\
10$^{4}$–10$^{7}$ & 40.1 & 20.8 & 15.6 & 15.4 \\
10$^{7}$–10$^{13}$ & 32.2 & 17.7 & 14.4 & 13.8 \\
10$^{13}$–10$^{31}$ & 18.9 & 14.3 & 12.8 & 12.0 \\
$\geq$10$^{31}$ & -- & 6.9 & 7.4 & 7.6 \\
\bottomrule
\end{tabular}
\end{table}

\begin{table}[H]
\small\centering\revcolor
\caption{Intermediate values of the reference execution by input range and data type: median $\lfloor\log_{10}|v|\rfloor$ of the largest and of the smallest non-zero intermediate value per example, and share of examples with a non-finite intermediate value.}
\label{tab:c5_reference_intermediate_stats_by_range_dtype}
\begin{tabular}{llrrr}
\toprule
Range & Type & Largest (median) & Smallest non-zero (median) & Non-finite (\%) \\
\midrule
$[0,1]$    & float & 1 & $-2$ & 3.9 \\
$[1,10]$   & float & 4 & $-2$ & 4.7 \\
$[1,10]$   & int   & 3 & $-2$ & 7.0 \\
$[10,100]$ & float & 9 & $-4$ & 6.6 \\
$[10,100]$ & int   & 9 & $-4$ & 7.0 \\
\bottomrule
\end{tabular}
\end{table}

\begin{table}[H]
\small\centering\revcolor
\caption{FAA with and without examples whose reference execution contains a non-finite intermediate value.}
\label{tab:c5_faa_nonfinite_sensitivity}
\begin{tabular}{llll}
\toprule
 & FAA all (\%) & FAA finite refs (\%) & Non-finite refs (\%) \\
\midrule
DS-1.5 & 3.9 & 4.0 & 5.8 \\
QW-4 & 38.6 & 38.9 & 5.0 \\
OL-7 & 2.0 & 1.7 & 5.8 \\
MI-7 & 1.5 & 1.4 & 5.8 \\
DS-7 & 19.7 & 20.2 & 5.7 \\
OT-7 & 49.0 & 48.6 & 5.7 \\
MN-14 & 46.4 & 46.0 & 5.8 \\
DS-14 & 31.0 & 31.3 & 5.8 \\
QW-14 & 42.9 & 42.9 & 5.8 \\
MG-24 & 47.2 & 46.9 & 5.8 \\
SV-30 & 12.1 & 12.3 & 5.8 \\
NM-30 & 33.8 & 33.6 & 5.8 \\
OA-120 & 60.0 & 60.3 & 5.8 \\
DS-685 & 81.5 & 81.6 & 0.1 \\
KK-1T & 98.7 & 98.7 & 0.0 \\
\bottomrule
\end{tabular}
\end{table}
\section{Comparison with Prior Work}
\label{app:prior_comparison}

\begin{table}[H]
\centering
\small
\begin{tabular}{lllll}
\toprule
\textbf{Work} & \textbf{Task} & \textbf{Procedure given} & \textbf{Complexity axis} & \textbf{Checked} \\
\midrule
\citet{shojaee2025illusion} & Puzzles & Partly & Problem size & Moves \\
\citet{varela2025rethinking} & Puzzles & Partly & Problem size & Moves \\
GSM-Symbolic~\citep{mirzadeh2025gsmsymbolic} & Word problems & No & Added clauses & Final answer \\
GSM-$\infty$~\citep{zhou2025gsminfty} & Word problems & No & Operations, context & Final answer \\
Ours & Arithmetic procedures & Yes, every step & Length, look-back & Every step \\
\bottomrule
\end{tabular}
\caption{Comparison with the closest prior work.}
\label{tab:prior_comparison}
\end{table}

\clearpage
\section{Results: Statistical Reliability of Accuracy Estimates}
\label{app:CI}

\begin{table}[H]
\centering
\small
\begin{tabular}{lc}
\toprule
\textbf{Model} & \textbf{FAA (\%)} \\
\midrule
Mistral-7B-Instruct-v0.3~\cite{mistral7b_instruct_v03} & $1.51 \pm 0.10$ \\
Olmo-3-7B-Think~\cite{olmo2025olmo} & $2.00 \pm 0.11$ \\
DeepSeek-R1-Distill-Qwen-1.5B~\cite{guo2025deepseek} & $3.94 \pm 0.16$ \\
Sarvam-30B~\cite{sarvam_sovereign_models} & $12.07 \pm 0.27$ \\
DeepSeek-R1-Distill-Qwen-7B~\cite{guo2025deepseek} & $19.71 \pm 0.33$ \\
DeepSeek-R1-Distill-Qwen-14B~\cite{guo2025deepseek} & $30.98 \pm 0.38$ \\
NVIDIA-Nemotron-3-Nano-30B~\cite{blakeman2025nvidia} & $33.77 \pm 0.40$ \\
Qwen3-4B-Thinking-2507~\cite{yang2025qwen3} & $38.64 \pm 0.41$ \\
Qwen3-14B~\cite{yang2025qwen3} & $42.91 \pm 0.39$ \\
Ministral-3-14B-Reasoning-2512~\cite{liu2026ministral} & $46.35 \pm 0.43$ \\
Magistral-Small-2509~\cite{rastogi2025magistral} & $47.21 \pm 0.40$ \\
OpenThinker2-7B~\cite{guha2025openthoughts} & $48.99 \pm 0.43$ \\
Kimi-K2.5~\cite{team2026kimi} & $56.32 \pm 0.39$ \\
DeepSeek-V3.2~\cite{deepseekai2025deepseekv32} & $56.58 \pm 0.41$ \\
gpt-oss-120b~\cite{agarwal2025gpt} & $60.04 \pm 0.42$ \\

\bottomrule
\end{tabular}
\caption{Model FAA with 95\% bootstrap confidence intervals. Values are reported as mean FAA ($\%$) $\pm$ confidence interval half-width. Confidence intervals quantify uncertainty arising from the evaluation examples only and do not reflect run-to-run variability, as all experiments use deterministic decoding (temperature = 0).}
\label{tab:model_ci}
\end{table}

\clearpage
\section{Model Generations}
\label{app:model_outputs}

\subsection{Example 1:}
\label{app:example_1}
\begin{tcolorbox}[breakable]

\textbf{Config:}
\begin{verbatim}
Range: 1 to 10
Input Type: Int
Task: Mixed
Look Back: 2
Id: "4179af72-0ce8-43b8-b92a-7273e8812f27"
\end{verbatim}

\textbf{Procedure:}
\begin{verbatim}
function(x, y):
Let S1 = x
Let S2 = y
Step 1: S3 = S1 + S2
Step 2: S4 = S3 - S2
Step 3: S5 = S4 - S2
Step 4: S6 = S4 - S2
Step 5: S7 = S2 + S5
Final Step: Return S7
\end{verbatim}

\textbf{Inputs:}
\begin{verbatim}
x = 8
y = 7
\end{verbatim}

\textbf{Expected Output:}
\begin{verbatim}
<answer>8</answer>
\end{verbatim}

\textbf{Correct Generation (OA-120):}
\begin{verbatim}
<answer>8</answer>
\end{verbatim}

\textbf{Incorrect Generation (OL-7):}
\begin{verbatim}
Now, execute the procedure with the given inputs.
<answer>42</answer>
\end{verbatim}

\textbf{Step-wise Correct Execution (OL-7):}
\begin{verbatim}
"Let's execute the procedure step by step.
Step 1: S1 = x
S1 = 8

Step 2: S2 = y
S2 = 7

Step 3: S3 = S1 + S2
S3 = 8 + 7 = 15

Step 4: S4 = S3 - S2
S4 = 15 - 7 = 8

Step 5: S5 = S4 - S2
S5 = 8 - 7 = 1

Step 6: S6 = S4 - S2
S6 = 8 - 7 = 1

Step 7: S7 = S2 + S5
S7 = 7 + 1 = 8

Final Step: Return S7\n<answer>8</answer>"
\end{verbatim}

\textbf{Step-wise Incorrect Execution (MI-7):}
\begin{verbatim}
"Output:
<answer>12</answer>

Explanation:
1. S1 = 8
2. S2 = 7
3. S3 = S1 + S2 = 8 + 7 = 15
4. S4 = S3 - S2 = 15 - 7 = 8
5. S5 = S4 - S2 = 8 - 7 = 1
6. S6 = S4 - S2 = 8 - 7 = 1 (This step is repeated, but the result is the same)
7. S7 = S2 + S5 = 7 + 1 = 8
8. Return S7 = 8

The final result is <answer>12</answer> (This is incorrect, as it
includes additional text)

Correct output:
S1 = 8
S2 = 7
S3 = 15
S4 = 8
S5 = 1
S6 = 1
S7 = 8
<answer>12</answer>"
\end{verbatim}

\label{fig:model_output_examples}

\end{tcolorbox}

\end{document}